\documentclass[10pt,twocolumn,letterpaper]{article}

\usepackage{cvpr}
\usepackage{times}
\usepackage{epsfig}
\usepackage{graphicx}
\usepackage{tabularx}
\usepackage{amsmath}
\usepackage{amssymb}
\usepackage{multirow}
\usepackage{subfigure}
\usepackage[noend]{algpseudocode}
\usepackage{algorithmicx,algorithm}

\usepackage[pagebackref=true,breaklinks=true,letterpaper=true,colorlinks,bookmarks=false]{hyperref}

\cvprfinalcopy 

\def\cvprPaperID{219} 

\newcommand{\R}[1]{\textcolor[rgb]{1.00,0.00,0.00}{#1}}
\newcommand{\B}[1]{\textcolor[rgb]{0.00,0.00,1.00}{#1}}
\makeatletter
\renewcommand{\@thesubfigure}{\hskip\subfiglabelskip}
\makeatother

\ifcvprfinal\fi
\begin{document}

\title{Blind Super-Resolution With Iterative Kernel Correction}

\author{Jinjin Gu$^{1}$\thanks{This work was done when they were interns at SenseTime.} , Hannan Lu$^{2\ast}$, Wangmeng Zuo$^2$, Chao Dong$^{3}$\\
$^1$The School of Science and Engineering, The Chinese University of Hong Kong, Shenzhen\\
$^2$School of Computer Science and Technology, Harbin Institute of Technology, Harbin, China\\
$^3$ShenZhen Key Lab of Computer Vision and Pattern Recognition, SIAT-SenseTime Joint Lab,\\Shenzhen Institutes of Advanced Technology, Chinese Academy of Sciences\\
{\tt\small jinjingu@link.cuhk.edu.cn, \{hannanlu, wmzuo\}@hit.edu.cn, chao.dong@siat.ac.cn}
}

\maketitle

\begin{abstract}
Deep learning based methods have dominated super-resolution (SR) field due to their remarkable performance in terms of effectiveness and efficiency.
Most of these methods assume that the blur kernel during downsampling is predefined/known (e.g., bicubic).
However, the blur kernels involved in real applications are complicated and unknown, resulting in severe performance drop for the advanced SR methods.
In this paper, we propose an Iterative Kernel Correction (IKC) method for blur kernel estimation in blind SR problem, where the blur kernels are unknown.
We draw the observation that kernel mismatch could bring regular artifacts (either over-sharpening or over-smoothing), which can be applied to correct inaccurate blur kernels.
Thus we introduce an iterative correction scheme -- IKC that achieves better results than direct kernel estimation.
We further propose an effective SR network architecture using spatial feature transform (SFT) layers to handle multiple blur kernels, named SFTMD.
Extensive experiments on synthetic and real-world images show that the proposed IKC method with SFTMD can provide visually favorable SR results and the state-of-the-art performance in blind SR problem.
\end{abstract}
\section{Introduction}
As a fundamental low-level vision problem, single image super-resolution (SISR) is an active research topic and has attracted increasingly attention.
SISR aims to reconstruct the high-resolution (HR) image from its low-resolution (LR) observation.
Since the seminal work of employing convolutional neural networks (CNNs) for SR \cite{SRCNN}, various deep learning based methods with different network architectures \cite{VDSR,DRCN,LapSRN,MemNet,DenseSR,haris2018deep,RCAN} and training strategies \cite{SRGAN,ESRGAN,ZSSR,bulat2018learn} have been proposed to continuously improve the SR performance.
Most of the existing advanced SR methods assume that the downsampling blur kernel is known and pre-defined, but the blur kernels involved in real applications are typically complicated and unavailable.
As has been revealed in \cite{efrat2013accurate,Benckmark}, learning-based methods will suffer severe performance drop when the pre-defined blur kernel is different from the real one.
This phenomenon of kernel mismatch will introduce undesired artifacts to output images, as shown in \figurename~\ref{fig:artifacts}. 
Thus the problem with unknown blur kernels, also known as \emph{blind SR}, has failed most of deep learning based SR methods and largely limited their usage in real-world applications.
\begin{figure}[!t]
	\begin{center}\vspace{-0.4cm}
		\subfigure[LR image]{\includegraphics[width=0.234\textwidth]{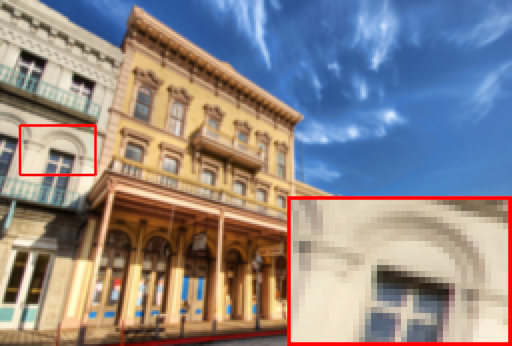}}
		\hspace{0.00cm}
		\subfigure[ZSSR \cite{ZSSR}]{\includegraphics[width=0.234\textwidth]{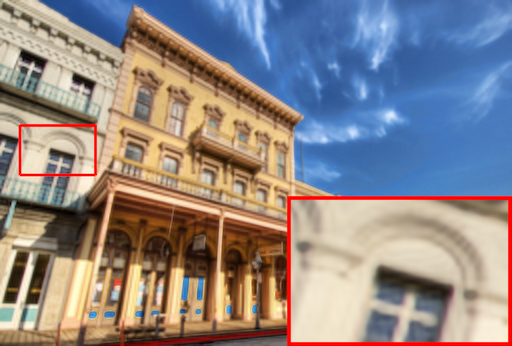}}

		\vspace{-0.23cm}
		\subfigure[SR without kernel correction]{\includegraphics[width=0.234\textwidth]{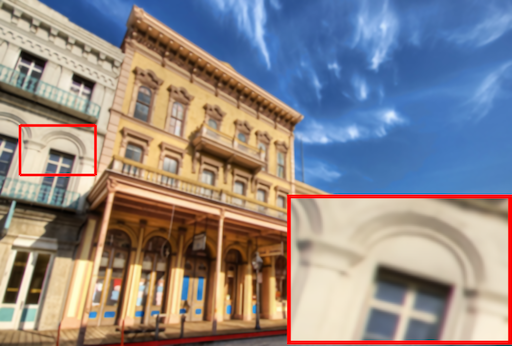}}
		\hspace{0.00cm}
		\subfigure[Iterative Kernel Correction (ours)]{\includegraphics[width=0.234\textwidth]{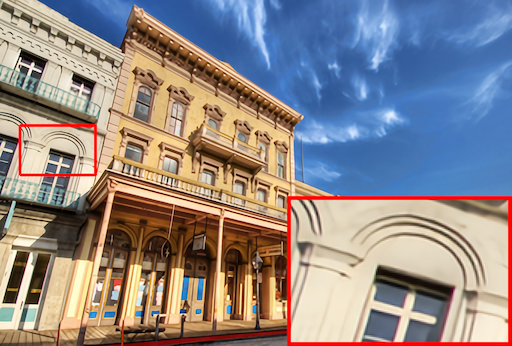}}
		\caption{SISR results on image ``\emph{img\_017}'' with SR factor 4. Before bicubic downsamping, the HR image is blurred by a Gaussian kernel with $\sigma=1.8$}
		\label{fig:open}
	\end{center}\vspace{-0.8cm}
\end{figure}

Most existing blind SR methods are model-based \cite{begin2004blind,wang2005patch,he2011single,he2009soft,joshi2008psf}, which usually involve complicated optimization procedures.
They predict the underlying blur kernel using self-similarity properties of natural images \cite{michaeli2013nonparametric}.
However, their predictions are easily affected by input noises, leading to inaccurate kernel estimation.
A few deep learning based methods have also tried to make progress for blind SR.
For example, in CAB \cite{CAB} and SRMD \cite{SRMD}, the network can take the blur kernel as an additional input and generate different results according to the provided kernel.
They achieve satisfactory performance if the input kernel is close to the ground truth.
However, these methods still cannot predict the blur kernel for every image on hand, thus are not applicable in real applications.
Although deep learning based methods have dominated SISR, they have limited progress on blind SR problem.

In this paper, we focus on using deep learning methods to solve the blind SR problem.
Our method stems from the observation that artifacts caused by kernel mismatch have regular patterns.
Specifically, if the input kernel is smoother than the real one, then the output image will be blurry/over-smoothing.
Conversely, if the input kernel is sharper than the correct one, then the results will be over-shapened with obvious ringing effects (see \figurename~\ref{fig:artifacts}).
This asymmetry of kernel mismatch effect provides us an empirical guidance on how to correct an inaccurate blur kernel.
In practical, we propose an Iterative Kernel Correction (IKC) method for blind SR based on predict-and-correct principle.
The estimated kernel is iteratively corrected by observing the previous SR results, and gradually approaches the ground truth.
Even the predicted blur kernel is slightly different from the real one, the output image can still get rid of those regular artifacts caused by kernel mismatch.

By further diving into the SR methods proposed for multiple blur kernels (i.e., SRMD \cite{SRMD}), we find that taking the concatenation of image and blur kernel as input is not the optimal choice.
To make a step forward, we employ spatial feature transform (SFT) layers \cite{SFTGAN} and propose an advanced CNN structure for multiple blur kernels, namely SFTMD.
Experiments demonstrate that the proposed SFTMD is superior to SRMD by a large margin. By combining the above components -- SFTMD and IKC, we achieve state-of-the-art (SOTA) performance on blind SR problem.

We summarize our contributions as follows:
(1) We propose an intuitive and effective deep learning framework for blur kernel estimation in single image super resolution. 
(2) We propose a new non-blind SR network using the spatial feature transform layers for multiple blur kernels.
We demonstrate the superior performance of the proposed non-blind SR network. 
(3) We test the blind SR performance on both carefully selected blur kernels and real images. Extensive experiments show that the combination of SFTMD and IKC achieves the SOTA performance in blind SR problem.
\section{Related Work}
\textbf{Super-Resolution Neural Networks.}
In the past few years, neural networks have shown remarkable capability on improving SISR performance.
Since the pioneer work of using CNN to learn the end-to-end mapping from LR to HR images \cite{SRCNN}, plenty of CNN architectures have been proposed for SISR \cite{FSRCNN,ESPCN,LapSRN,haris2018deep,DRCN,DRRN}.
In order to go deeper in network depth and achieve better performance, most of the existing high-performance SR networks have residual architecture \cite{VDSR}.
SRGAN \cite{SRGAN} first introduce residual blocks into SR networks.
EDSR \cite{EDSR} improve it by removing unnecessary batch normalization layer in residual block and expanding the model size.
DenseSR \cite{DenseSR} present an effective residual dense block and ESRGAN \cite{ESRGAN} further use a residual-in-residual dense block to improve the perceptual quality of SR results.
Zhang \textit{et al.} \cite{RCAN} introduce the channel attention component in residual blocks.
Some networks are specifically designed for the SR task in some special scenarios, e.g., Wang \textit{et al.} \cite{SFTGAN} use a novel spatial feature transform layer to introduce the semantic prior as an additional input of SR network.
Moreover, Riegler \textit{et al.} \cite{CAB} propose conditioned regression models can effectively exploit the additional kernel information during training and inference.
SRMD \cite{SRMD} propose a stretching strategy to integrate non-image degradation information in a SR network.

\textbf{Blind Super-Resolution.}
Blind SR assume that the degradation kernels are unavailable.
In recent years, the community has paid relatively less research attention to blind SR problem.
Michaeli and Irani \cite{michaeli2013nonparametric} estimate the optimal blur kernel based on the property that small image patches  will re-appear in images.
There are also research works trying to employ deep learning in blind SR task.
Yuan \textit{et al.} \cite{yuan2018unsupervised} propose to learn not only SR mapping but also the degradation mapping using unsupervised learning.
Shocher \textit{et al.} \cite{ZSSR} exploit the internal recurrence of information inside an image and propose an unsupervised SR method to super-resolve images with different blur kernels.
They train a small CNN on examples extracted from the input image itself, the trained image-specific CNN is appropriate for super-resolving this image.
Different from the previous works, our method employs the correlation between SR results and kernel mismatch.
Our method uses the intermediate SR results to iteratively correct the estimation of blur kernels, thus provide artifact-free final SR results.
\section{Method}
\subsection{Problem Formulation}
The blind super-resolution problem is formulated as follows.
Mathematically, the HR image $I^{HR}$ and LR image $I^{LR}$ are related by a degradation model
\begin{equation}
	I^{LR}=(k\otimes I^{HR})\downarrow_s+n,
	\label{eq:degrad}
\end{equation}
where $\otimes$ denotes convolution operation.
There are three main components in this model, namely the blur kernel $k$, the downsampling operation $\downarrow_s$ and the additive noise $n$.
In literature, the most widely adopted blur kernel is isotropic Gaussian blur kernel \cite{dong2013nonlocally,Benckmark,SRMD}.
Besides, the anisotropic blur kernels also appear in some works \cite{CAB,SRMD}, which can be regarded as the combination of a motion blur and an isotropic blur kernel.
For simplicity, we mainly focus on the isotropic blur kernel without motion effect in this paper.
Following most recent deep learning based SR methods \cite{SRMD}, we adopt the combination of Gaussian blur and bicubic downsampling.
In real-world use cases, the LR images are often accompanied with additive noises.
As in SRMD \cite{SRMD}, we assume that the additive noise follows Gaussian distribution in real world application.
Note that the formulation of blind SR in this paper is different with the previous works \cite{michaeli2013nonparametric,yuan2018unsupervised} .
Although defined as blind SR problem, our method focuses on a limited variety of kernels and noise.
But the kernel estimated according to our assumptions can handle most of the real world images.
\subsection{Motivation}
We then review the importance of using correct blur kernel during SISR based on the settings described above.
In order to obtain the LR images $I^{LR}$, the HR images $I^{HR}$ are first blurred by the isotropic Gaussian kernel with kernel width $\sigma_{LR}$ and then downsampled by bicubic interpolation.
Assume that the mapping $\mathcal{F}(I^{LR}, k)$ is a well-trained SR model with the kernel information as input (e.g., SRMD \cite{SRMD}).
Then the output image is artifact-free with correct kernel $k$.
The blind SR problem is equivalent to finding the kernel $k$ that helps SR model generate visual pleasing result $I^{SR}$.
A straightforward solution is to adopt a predictor function $k'=\mathcal{P}(I^{LR})$ that estimates $k$ from the LR input directly.
The predictor can be optimized by minimizing the $l_2$ distance as
\begin{equation}
	\theta_\mathcal{P}=\mathop{\arg\min}_{\theta_\mathcal{P}}\|k-\mathcal{P}(I^{LR};\theta_\mathcal{P})\|_2^2,
	\label{eq:lossp}
\end{equation}
where $\theta_\mathcal{P}$ is the parameter of $\mathcal{P}$.
By employing the predictor function and the SR model together, we are able to build an end-to-end blind SR model.

\begin{figure}[!t]
	\centering
	\includegraphics[width=0.94\linewidth]{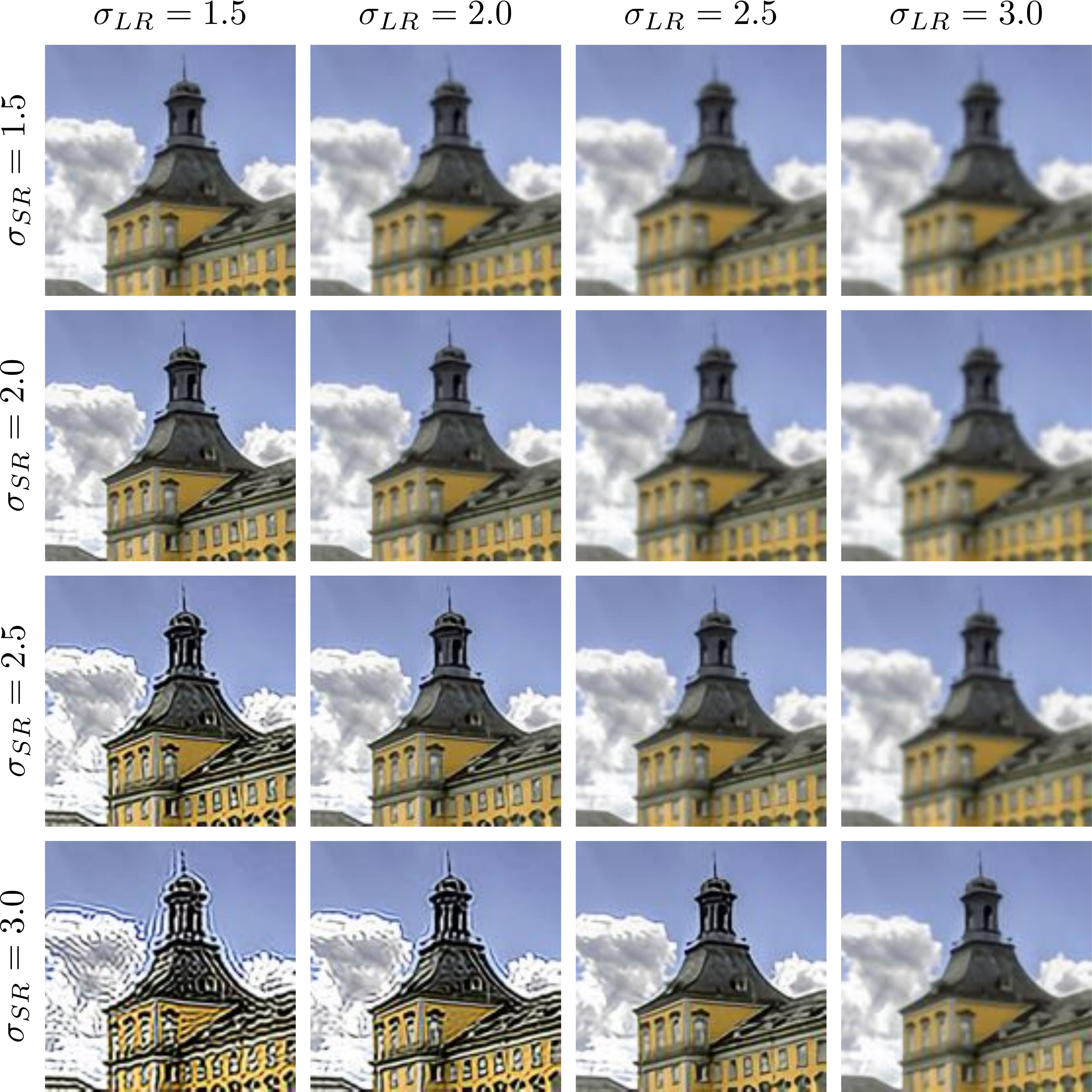}
	\caption{SR sensitivity to the kernel mismatch. Where $\sigma_{LR}$ denotes the kernel used for downsampling and $\sigma_{SR}$ denotes the kernel used for SR.}
	\label{fig:artifacts}
	\vspace{-0.4cm}
\end{figure}

However, accurate estimation of $k$ is impossible.
As the inverse problem is ill-posed, there exists multiple candidates of $k$ for a single input.
Meanwhile, the SR models are very sensitive to the estimation error.
If the inaccurate kernel is used for SR directly, then the final SR results will contain obvious artifacts.
\figurename~\ref{fig:artifacts} shows the sensitivity of the SR results to kernel mismatch, where $\sigma_{SR}$ denotes the kernel width used for SR.
As shown in the upper-right region of \figurename~\ref{fig:artifacts}, where the kernel used for SR are sharper than the real one ($\sigma_{SR}<\sigma_{LR}$), the SR results are over-smoothing and the the high frequency textures are significantly blurred.
In the lower-left region of \figurename~\ref{fig:artifacts}, where the kernel used for SR are smoother than the correct one ($\sigma_{SR}>\sigma_{LR}$), the SR results show unnatural ringing artifacts caused by over-enhancing high-frequency edges.
In contrast, the results on the diagonal, which use correct blur kernels, look natural without artifacts and blurring.
The above phenomenon illustrates that the estimation error of $k$ will be significantly magnified by the SR model, resulting in unnatural output images.
To address the kernel mismatch problem, we propose to iteratively correct the kernel until we obtain an artifact-free SR results.

To correctly estimate $k$, we build a corrector function $\mathcal{C}$ that measures the difference between the estimated kernel and the ground truth kernel.
In the core of our idea is to adopt the intermediate SR results.
The corrector function can be obtained by minimizing the $l_2$ distance between the corrected kernel and the ground truth as
\begin{equation}
	\theta_\mathcal{C}=\mathop{\arg\min}_{\theta_\mathcal{C}}\|k-(\mathcal{C}(I^{SR};\theta_\mathcal{C})+k')\|_2^2,
	\label{eq:lossc}
\end{equation}
where $\theta_\mathcal{C}$ is the parameter of $\mathcal{C}$ and $I^{SR}$ is the SR result using the last estimated kernel.
This corrector adjusts the estimated blur kernel based on the features of the SR image.
After correction, the SR results using adjusted kernel are supposed to approach natural images with less artifacts.

\begin{figure}[t]
	\centering
	\setlength{\tabcolsep}{0.5mm}
	\begin{tabular}{cc}
		\hspace{-0.1cm}\includegraphics[width=0.47\linewidth]{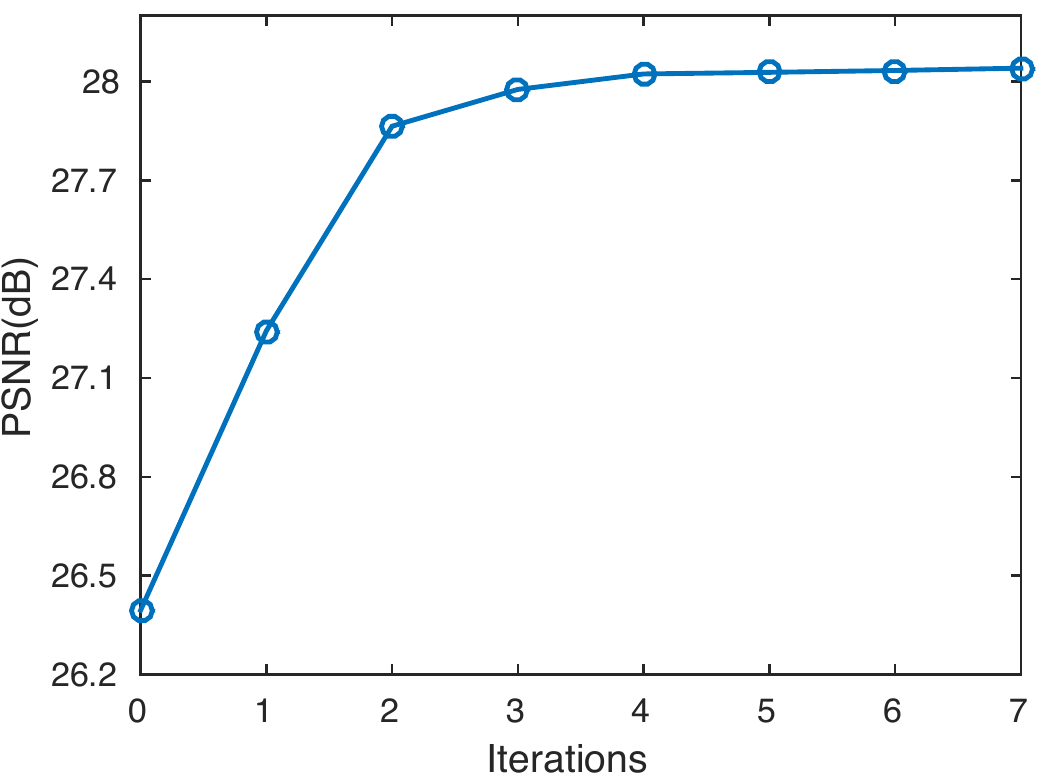}&
		\includegraphics[width=0.48\linewidth]{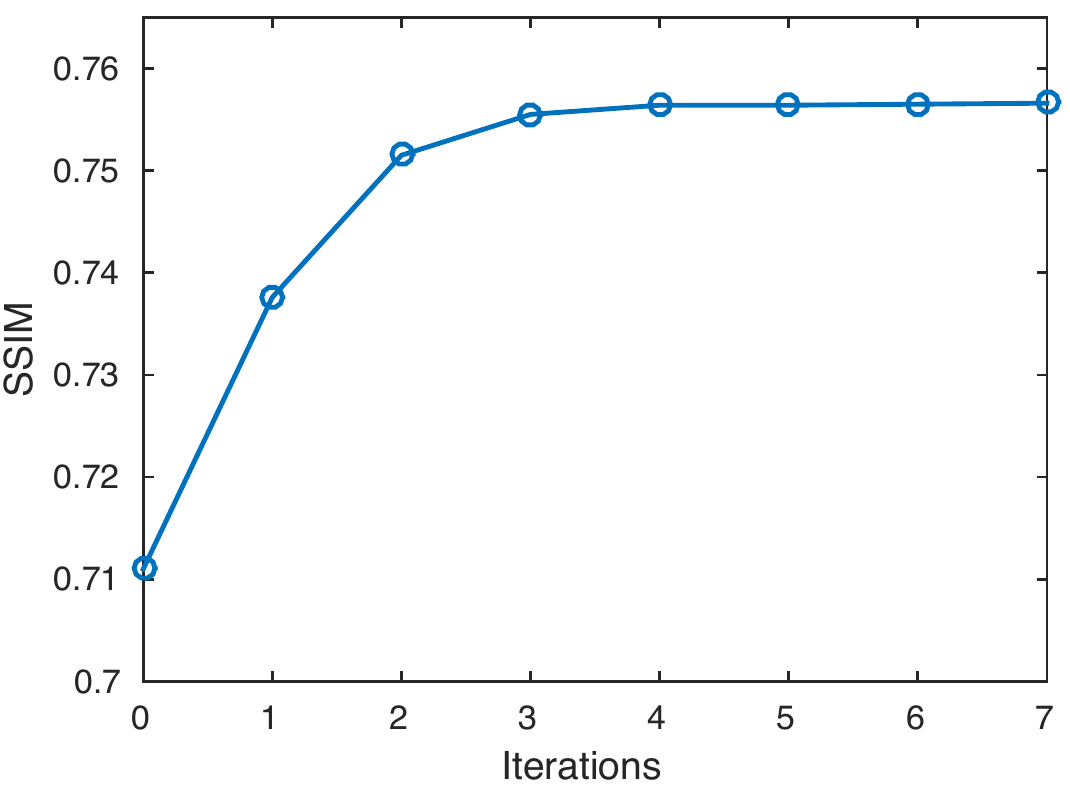}
	\end{tabular}
	\caption{The curves of PSNR and SSIM vs. iterations. The experiments are conducted using IKC method. The test set is Set14 and the SR factor is 4.}
	\label{fig:steps}
	\vspace{-0.4cm}
\end{figure}

However, if we train our model with only one time of correction, the corrector may provide inadequate correction or over-correct the kernel, leading to unsatisfactory SR results.
A possible solution is to use smaller correction steps that gradually refine the kernel until it reaches ground truth.
When the SR result does not contain serious over-smoothing or over-sharpening effects, the corrector will make little changes to the estimated kernel to ensure convergence.
Then we are able to get a high-quality SR image by iteratively applying kernel correction.
Experiments also demonstrate our assumption.
\figurename~\ref{fig:steps} shows the PSNR and SSIM results using different iteration numbers.
It can be observed that correcting only once is not sufficient.
When the number of iterations increases, both PSNR and SSIM increase gradually until convergence.
\subsection{Proposed Method}
\begin{figure*}[ht]
	\begin{center}\vspace{-0.2cm}
	\includegraphics[width=1.0\linewidth]{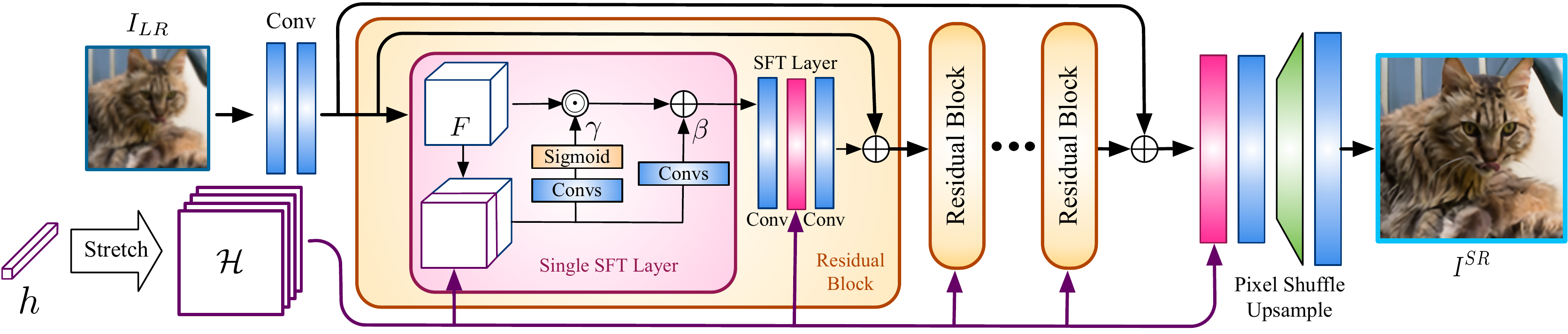}
	\caption{The architecture of the proposed SFTMD network. The design of the proposed SFT layer is shown in pink box.}
	\label{fig:SFTMD}
	\end{center}\vspace{-0.2cm}
\end{figure*}
\begin{figure*}
	\begin{center}\vspace{-0.4cm}
	\includegraphics[width=0.9\linewidth]{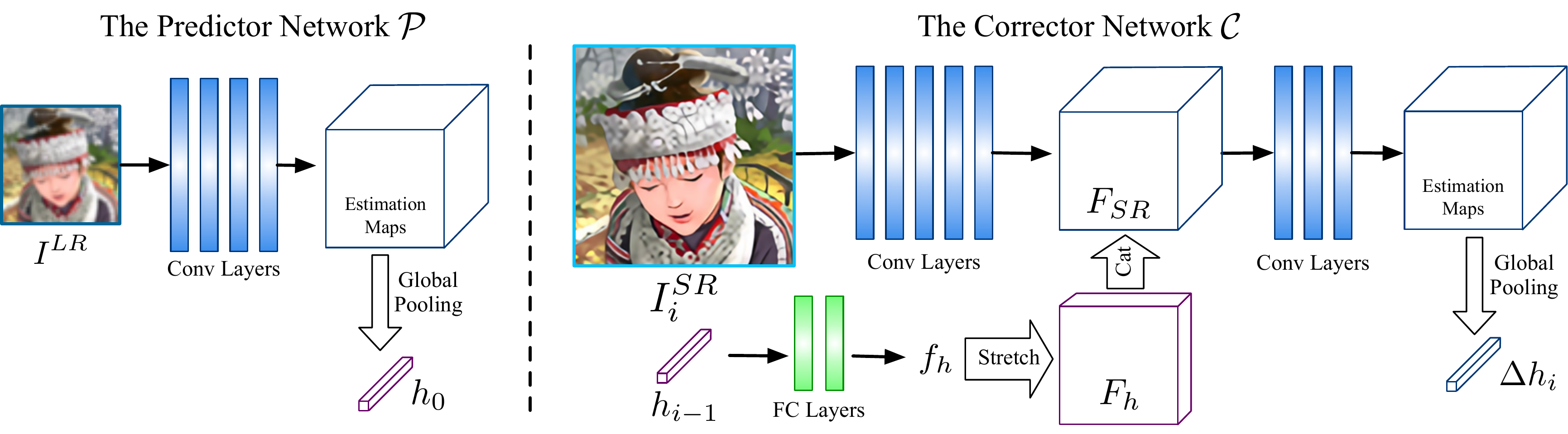}
	\caption{The network architecture of the proposed predictor and corrector.}
	\label{fig:Network}
	\end{center}\vspace{-0.8cm}
\end{figure*}
\textbf{Overall framework.}
The proposed Iterative Kernel Correction (IKC) framework consists of a SR model $\mathcal{F}$, a predictor $\mathcal{P}$ and a corrector $\mathcal{C}$, and the pseudo-code is shown in Algorithm 1.
Suppose the LR image $I^{LR}$ is of size $C\times H\times W$, where $C$ denotes the number of channels, $H$ and $W$ denote the height and width of the image.
We assume that blur kernel is of size $l\times l$ and the kernel space is a $l^2$-dimensional linear space.
In order to save computation, we first reduce the dimensionality of the kernel space by principal component analysis (PCA).
The kernels are projected onto a $b$-dimensional linear space by a dimension reduction matrix $M\in\mathbb{R}^{b\times l^2}$.
Thus we only need to perform estimation in this low dimensional space, which is more effective in calculation.
The kernel after the dimension reduction is denoted by $h$, where $h=Mk,h\in\mathbb{R}^b$.
At the start of the algorithm, an initial estimation $h_0$ is given by the predictor function $h_0=\mathcal{P}(I^{LR})$, and then used to get the first SR result $I_0^{SR}=\mathcal{F}(I^{LR},h_0)$.
After obtaining the initial estimation, we proceed to the correction phase of the estimated kernel.
At the $i$th iteration, given the previous estimation $h_{i-1}$, the correcting update $\Delta h_i$, the new estimation $h_i$ and the new SR result $I_i^{SR}$ can be written as
\begin{align}
	\Delta h_i&=\mathcal{C}(I^{SR}_i, h_{i-1})\\
	h_i&=h_{i-1}+\Delta h_i\\
	I^{SR}_i&=\mathcal{F}(I^{LR}, h_i).
\end{align}
After $t$ iterations, the $I^{SR}_t$ is the final output of IKC.

\textbf{Network architecture of SR model $\mathcal{F}$.}
As the most successful SR method for multiple blur kernels, SRMD \cite{SRMD} propose a simple yet efficient stretching strategy for CNN to process non-image input directly.
SRMD stretches the input $h$ into kernel maps $\mathcal{H}$ of size $b\times H\times W$, where all the elements of the $i$th map are equal to the $i$th element of $h$.
SRMD takes the concatenated LR image and kernel maps of size $(b+C)\times H\times W$ as input.
Then, a cascade of $3\times 3$ convolution layers and one pixel-shuffle upsampling layer are applied to perform super-resolution.
However, to exploit the kernel information, concatenating the image and the transformed kernel as input is not the only or best choice.
On the one hand, the kernel maps do not actually contain the information of the image.
Processing the kernel maps and the image at the same time with convolution operation will introduce interference that is not related to the image.
Using this concatenation strategy with residual blocks can interfere with image processing, making it difficult to employ residual structure to improve performance.
On the other hand, the influence of kernel information is only considered at the first layer.
When applying the same strategy in a deeper network, the deeper layers are difficult to be affected by the kernel information input at the first layer.
To address above problems, we proposed a new SR model for multiple kernels using spatial feature transform (SFT) layers \cite{SFTGAN}, namely SFTMD.
In SFTMD, the kernel maps influence the output of network by applying an affine transformation to the feature maps in each middle layer by SFT layers.
This affine transformation is not involved in the process of input image directly, thus providing better performance.
\begin{algorithm}[t]
\caption{Iterative Kernel Correction}
\hspace*{0.02in} {\bf Require:}
the LR image $I^{LR}$\\
\hspace*{0.02in} {\bf Require:}
the max iteration number $t$
\begin{algorithmic}[1]
\State $h_0\gets\mathcal{P}(I^{LR})$ (Initialize the kernel estimation)
\State $I^{SR}_0\gets\mathcal{F}(I^{LR},h_0)$ (The initial SR result)
\State $i\gets0$ (Initialize counter)
\While{$i<t$}
　　\State $i\gets i+1$
　　\State $\Delta h_i\gets\mathcal{C}(I^{SR}_{i-1}, h_{i-1})$ (Estimate the kernel error using the intermediate SR results)
　　\State $h_i\gets h_{i-1}+\Delta h_i$ (Update kernel estimation)
　　\State $I^{SR}_i\gets \mathcal{F}(I^{LR},h_i)$ (Update the SR result)
\EndWhile
\State \Return $I^{SR}_t$ (Output the final SR result)
\end{algorithmic}
\end{algorithm}

\figurename~\ref{fig:SFTMD} illustrates the network architecture of SFTMD.
We employ the high level architecture of SRResNet \cite{SRGAN} and extend it to handle multiple kernels by SFT layers.
The SFT layer provides affine transformation for the feature maps $F$ conditioned on the kernel maps $\mathcal{H}$ by a scaling and shifting operation:
\begin{equation}
	\mathrm{SFT}(F,\mathcal{H})=\gamma\odot F+\beta,
\end{equation}
where $\gamma$ and $\beta$ is the parameters for scaling and shifting, $\odot$ present Hadamard product.
The transformation parameters $\gamma$ and $\beta$ are obtained by small CNN.
Suppose that the output feature maps of the previous layer $F$ are of size $C_f \times H \times W$, where $C_f$ is the number of feature maps, and the kernel maps are of size $b \times H\times W$.
The CNN takes the concatenated feature maps and kernel maps (total size is $(b+C_f)\times H\times W$) as input and output $\gamma$ and $\beta$.
We use SFT layers after all convolution layers in residual blocks and after the global residual connection.
It is worth pointing out that the code maps are spatially uniform, thus the SFT layers do not actually provide spatial variability according to the code maps.
This is different from its application in semantic super resolution \cite{SFTGAN}.
We only employ the transformation characteristic of SFT layers.

\begin{table*}[ht]
	\scriptsize
	\begin{center}\vspace{-0.4cm}
	\caption{Quantitative comparison of SRCNN-CAB \cite{CAB}, SRMDNF \cite{SRMD} and the proposed SFTMD. The comparison is conducted using three different isotropic Gaussian kernels on Set5, Set14 and BSD100 dataset. The best two results are highlighted in \textcolor[rgb]{1.00,0.00,0.00}{red} and \textcolor[rgb]{0.00,0.00,1.00}{blue} colors.}
	\label{tab:SFT}
	\vspace{0.8mm}
	\begin{tabular}{p{5.4cm}p{1.6cm}<{\centering}ccccccccc}
	\hline
	\multirow{2}{*}{Method} & \multirow{2}{*}{Kernel Width} & \multicolumn{3}{c}{Set5 \cite{Set5}} &  \multicolumn{3}{c}{Set14 \cite{Set14}} &  \multicolumn{3}{c}{BSD100 \cite{BSD100}}\\
	&  & $\times 2$ & $\times 3$ & $\times 4$ & $\times 2$ & $\times 3$ & $\times 4$ & $\times 2$ & $\times 3$ & $\times 4$ \\
	\hline
	\hline
	SRCNN-CAB \cite{CAB} & \multirow{5}{*}{$0.2$} 
	& 33.27 & 31.03 & 29.31
	& 30.29 & 28.29 & 26.91 
	& 28.98 & 27.65 & 25.51 \\
	SRMDNF \cite{SRMD} &
	& \B{37.79} & \B{34.13} & \B{31.96}
	& \B{33.33} & \B{30.04} & \B{28.35} 
	& \B{32.05} & \B{28.97} & \B{27.49}\\
	SRResNet, concatenate at the first layer & 
	& 31.74 & 30.90 & 29.40 
	& 27.57 & 26.40 & 26.18 
	& 27.24 & 26.43 & 26.34 
	\\
	SRResNet, replace SFT layer by direct concatenation& 
	& 37.69 & 34.01 & 31.64  
	& 33.26 & \B{30.04} & 28.23  
	& 31.83 & 28.81 & 27.26  
	\\
	SFTMD (ours) & 
	&\R{38.00}&\R{34.57}&\R{32.39} 
	&\R{33.68}&\R{30.47}&\R{28.77} 
	&\R{32.09}&\R{29.09}&\R{27.58} 
	\\
	\hline
	SRCNN-CAB \cite{CAB} & \multirow{5}{*}{$1.3$} 
	& 33.42 & 31.14 & 29.50
	& 30.51 & 28.34 & 27.02 
	& 29.02 & 27.91 & 25.66 \\
	SRMDNF \cite{SRMD} & 
	& \B{37.44} & \B{34.17} & \B{32.00}
	& \B{33.20} & 30.08 & \B{28.42} 
	& \B{31.98} & \B{29.03} & \B{27.53}\\
	SRResNet, concatenate at the first layer &
	& 30.88 & 30.33 & 29.11 
	& 27.16 & 25.84 & 25.93 
	& 26.84 & 25.92 & 26.20 
	\\
	SRResNet, replace SFT layer by direct concatenation & 
	& 37.01 & 34.02 & 31.69 
	& 32.96 & \B{30.13} & 28.29 
	& 31.58 & 28.89 & 27.29 
	\\
	SFTMD (ours) & 
	&\R{37.46}&\R{34.53}&\R{32.41} 
	&\R{33.39}&\R{30.55}&\R{28.82} 
	&\R{32.06}&\R{29.15}&\R{27.64} 
	\\
	\hline
	SRCNN-CAB \cite{CAB} & \multirow{5}{*}{$2.6$} 
	& 32.21 & 30.82 & 28.81
	& 29.74 & 27.83 & 26.15 
	& 28.35 & 26.63 & 25.13 \\
	SRMDNF \cite{SRMD} &
	& \B{34.12} & \B{33.02} & \B{31.77}
	& \B{30.25} & \B{29.33} & \B{28.26} 
	& \B{29.23} & \B{28.35} & \B{27.43}\\
	SRResNet, concatenate at the first layer &
	& 24.22 & 28.44 & 28.64 
	& 22.99 & 24.19 & 25.63 
	& 23.07 & 24.42 & 25.99 
	\\
	SRResNet, replace SFT layer by direct concatenation & 
	& 27.75 & 32.71 & 31.35 
	& 25.67 & 29.28 & 28.07 
	& 25.57 & 28.19 & 27.15 
	\\
	SFTMD (ours) & 
	&\R{34.27}&\R{33.22}&\R{32.05} 
	&\R{30.38}&\R{29.63}&\R{28.55} 
	&\R{29.35}&\R{28.41}&\R{27.47} 
	\\
	\hline
	\end{tabular}
	\end{center}\vspace{-0.6cm}
\end{table*}
\textbf{Network architecture of predictor $\mathcal{P}$ and corrector $\mathcal{C}$.}
The network designs of the predictor and corrector are shown in \figurename~\ref{fig:Network}.
For the predictor $\mathcal{P}$, we use four convolution layers with Leaky ReLU activations and a global average pooling layer.
The convolution layers give the estimation of the kernel $h$ spatially and form the estimation maps. 
Then the global average pooling layer gives the global estimation by taking the mean value spatially.

For the corrector $\mathcal{C}$, we take not only the SR image $I^{SR}$ but also the previous estimation $h$ as inputs.
Similar to Eq. \eqref{eq:lossc}, the new corrector can be obtained by solving the following optimization problem:
\begin{equation}
	\theta_\mathcal{C}=\mathop{\arg\min}_{\theta_\mathcal{C}}\|k-(\mathcal{C}(I^{SR},h;\theta_\mathcal{C})+k')\|_2^2.
	\label{eq:lossc2}
\end{equation}
The input SR result is first processed to feature maps $F_{SR}$ by five convolution layers with Leaky ReLU activations.
Note that the previous SR result may contain artifacts (e.g., ringing and blurry) caused by kernel mismatch, which can be extracted by these convolution layers.
At the same time, we use two fully-connected layers with Leaky ReLU activations to extract the inner correlations of the previous kernel estimation.
We then stretch the output vector $f_h$ to feature maps $F_{h}$ using the same strategy used in SFTMD.
The $F_h$ and $F_{SR}$ are then concatenated to predict the $\Delta h$.
We use three convolution layers with kernel size $1\times 1$ and Leaky ReLU activations to give the estimation for $\Delta h$ spatially.
Same as the predictor, a global average pooling operation is used to get the global estimation of $\Delta h$.
\section{Experiments}
\subsection{Data Preparation and Network Training}
We synthesize the training image pairs according to the problem formulation described in section 3.1.
For the isotropic Gaussian blur kernels used for training, the kernel width ranges are set to $[0.2,2.0]$, $[0.2,3.0]$ and $[0.2,4.0]$ for SR factors $2$, $3$ and $4$, respectively.
We uniformly sample the kernel width in the above ranges.
The kernel size is fixed to $21\times 21$.
When applying on real world images, we use the additive Gaussian noise with covariance $\sigma=15$.
We also provide noise-free version for comparison on the synthetic test images.
The  HR images are collected from DIV2K \cite{DIV2K} and Flickr2K \cite{Flickr2K}, then the training set consists of $3450$ high-quality 2K images.
The training dataset is augmented with random horizontal flips and 90 degree rotations.
All models are trained and tested on RGB channels.

The SFTMD and IKC are both trained on the synthetic training image pairs and their corresponding blur kernels.
First, the SFTMD is pre-trained using mean square error (MSE) loss.
We then train the predictor network and the corrector network alternately.
The parameters of the trained SFTMD are fixed during training the predictor and the corrector.
The order of training can refer to Algorithm 1.
For every mini-batch data $\{I^{LR}_i, I^{HR}_i, h_i\}_{i=1}^N$, where $N$ denotes the mini-batch size, we first update the parameters of the predictor according to Eq. \eqref{eq:lossp}.
We then update the corrector according to Eq. \eqref{eq:lossc2} with a fixed iteration number $t=7$.
For optimization, we use Adam \cite{Adam} with $\beta_1=0.9$, $\beta_2=0.999$ and learning rate $1\times 10^{-4}$.
We implement our models with the PyTorch framework and train them using NVIDIA Titan Xp GPUs.

We also propose a test kernel set for the quantitative evaluation of blind SR methods, namely \textit{Gausssian8}.
As declared by the name, \textit{Gausssian8} consists eight selected isotropic Gaussian blur kernels for each SR factor $2$, $3$ and $4$ (twenty four kernels in total).
The ranges of kernel width are set to $[0.80,1.60]$, $[1.35,2.40]$ and $[1.80,3.20]$ for SR factors $2$, $3$ and $4$, respectively.
The HR images are first blurred by the selected blur kernels and then downsampled by bicubic interpolation.
By determining the blur kernels for testing, we can compare and analyze the performance of blind SR methods.
Although it only contains isotropic Gaussian kernels, it can still be used to test the basic performance of a blind SR method.
\begin{figure*}[!t]\vspace{-0.3cm}
	\scriptsize{
	\begin{center}
		\subfigure[A+ \cite{A+}]{\includegraphics[width=0.16\linewidth]{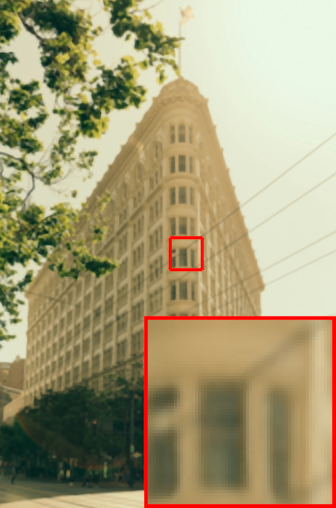}}
		\hspace{0.00cm}
		\subfigure[CARN \cite{CARN}]{\includegraphics[width=0.16\linewidth]{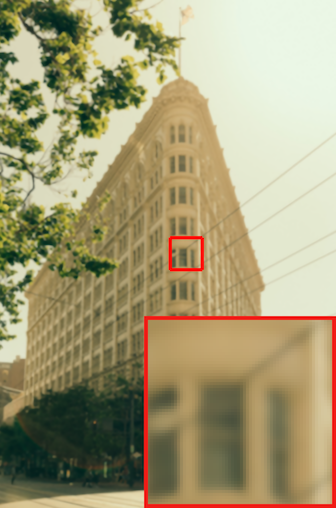}}
		\hspace{0.00cm}
		\subfigure[CARN $+$ Pan \textit{et al.}\cite{Deblur}]{\includegraphics[width=0.16\linewidth]{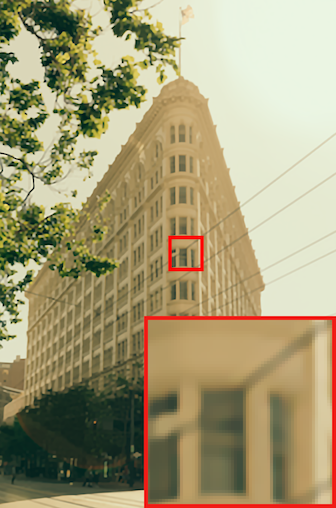}}
		\hspace{0.00cm}
		\subfigure[ZSSR \cite{ZSSR}]{\includegraphics[width=0.16\linewidth]{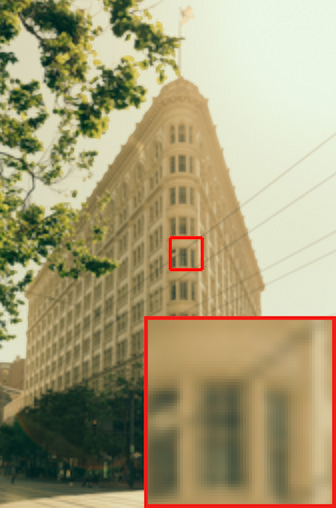}}
		\hspace{0.00cm}
		\subfigure[$\mathcal{P}+$SFTMD]{\includegraphics[width=0.16\linewidth]{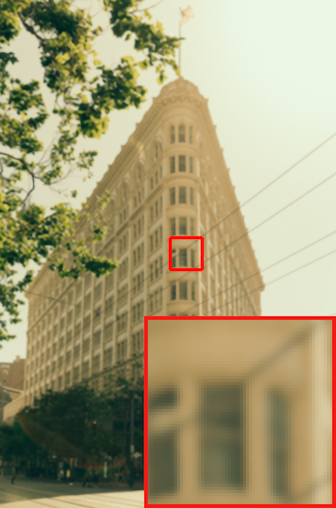}}
		\hspace{0.00cm}
		\subfigure[IKC (ours)]{\includegraphics[width=0.16\textwidth]{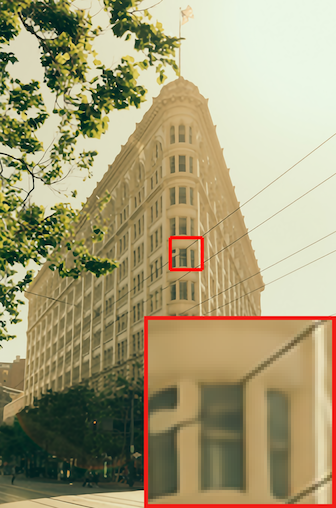}}
		\caption{SISR performance comparison of different methods with SR factor 4 and kernel width 1.8 on image ``\emph{Img\_050}'' from Urban100.}
		\label{fig:Gaus}\vspace{-0.6cm}
	\end{center}}
\end{figure*}
\begin{figure}[!t]
	\begin{center}
	\includegraphics[width=0.96\linewidth]{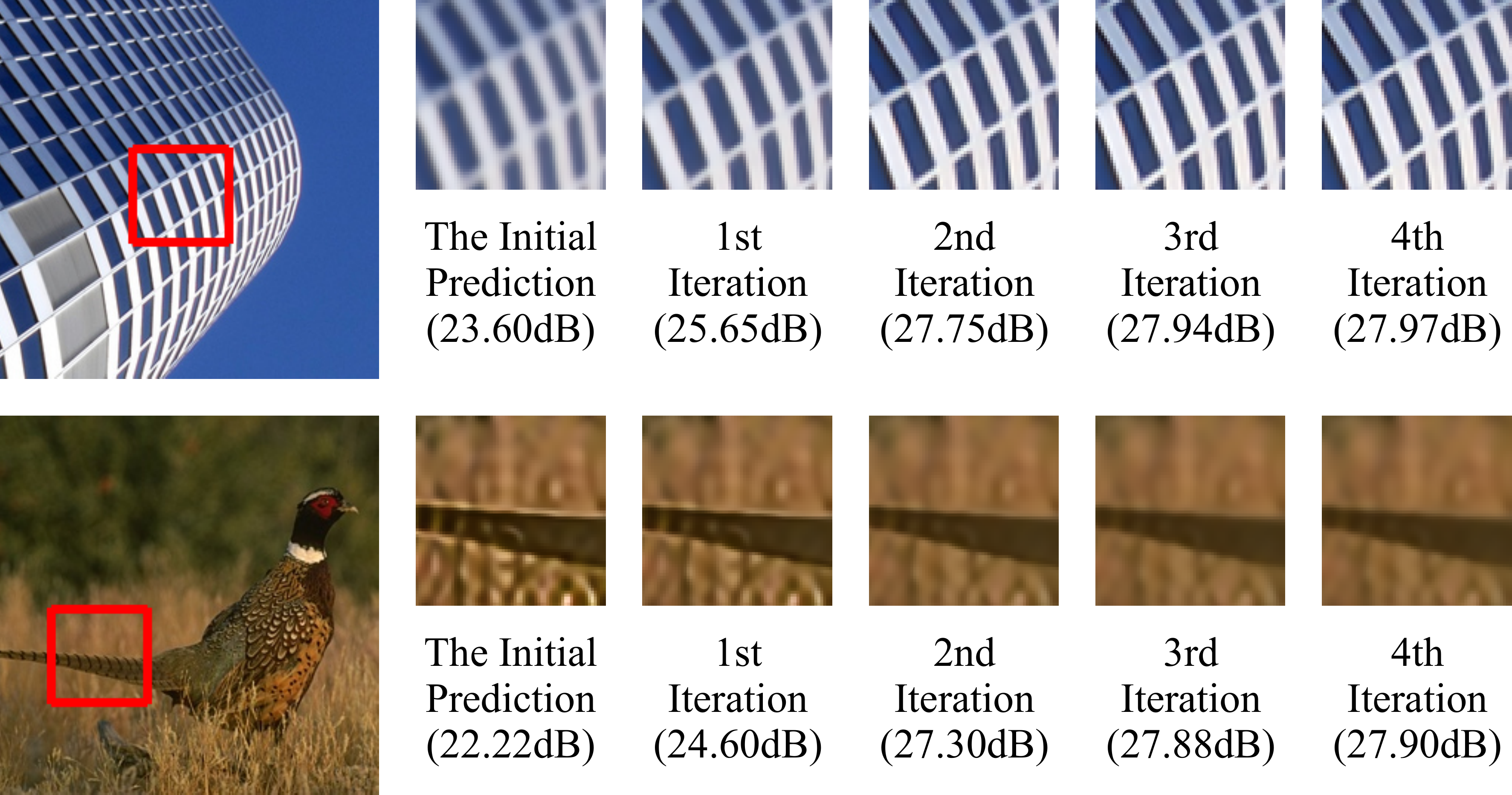}
	\caption{The intermediate SR results during kernel correction.}
	\label{fig:stepcom}
	\end{center}\vspace{-0.8cm}
\end{figure}
\begin{table*}[ht]
	\scriptsize
	\begin{center}
	\caption{Quantitative comparison of the SOTA SR methods and IKC method. The best two results are highlighted in \textcolor[rgb]{1.00,0.00,0.00}{red} and \textcolor[rgb]{0.00,0.00,1.00}{blue} colors, respectively. Note that the methods marked with ``*'' is not designed for blind SR, thus the comparison with these methods is unfair.}
	\label{tab:Gaus8}
	\vspace{0.8mm}
	\begin{tabular}{lp{1.6cm}<{\centering}cccccccccc}
	\hline
	\multirow{2}{*}{Method} & \multirow{2}{*}{Scale} &  \multicolumn{2}{c}{Set5 \cite{Set5}} &  \multicolumn{2}{c}{Set14 \cite{Set14}} &  \multicolumn{2}{c}{BSD100 \cite{BSD100}} &  \multicolumn{2}{c}{Urban100 \cite{Urban100}} &  \multicolumn{2}{c}{Manga109 \cite{Manga109}}\\
	&  & PSNR & SSIM & PSNR & SSIM & PSNR & SSIM & PSNR & SSIM & PSNR & SSIM \\
	\hline
	\hline
	Bicubic & \multirow{7}{*}{$\times 2$} 
	& 28.82 & 0.8577 
	& 26.02 & 0.7634 
	& 25.92 & 0.7310 
	& 23.14 & 0.7258 
	& 25.60 & 0.8498\\
	CARN$^*$ \cite{CARN} &
	& 30.99 & 0.8779 
	& 28.10 & 0.7879 
	& 26.78 & 0.7286 
	& 25.27 & 0.7630 
	& 26.86 & 0.8606\\
	ZSSR \cite{ZSSR} &  
	& 31.08 & 0.8786 
	& 28.35 & 0.7933 
	& 27.92 & 0.7632 
	& 25.25 & 0.7618 
	& 28.05 & 0.8769\\
	Pan \textit{et al.} \cite{Deblur} $+$ CARN \cite{CARN} &  
	& 24.20 & 0.7496 
	& 21.12 & 0.6170 
	& 22.69 & 0.6471 
	& 18.89 & 0.5895 
	& 21.54 & 0.7496\\
	CARN \cite{CARN} $+$ Pan \textit{et al.} \cite{Deblur} &  
	& 31.27 & 0.8974 
	& 29.03 & 0.8267 
	& 28.72 & 0.8033 
	& 25.62 & 0.7981 
	& 29.58 & 0.9134\\
	$\mathcal{P}+$ SFTMD &  
	& \textcolor[rgb]{0.00,0.00,1.00}{35.44} & \textcolor[rgb]{0.00,0.00,1.00}{0.9617} 
	& \textcolor[rgb]{0.00,0.00,1.00}{31.27} & \textcolor[rgb]{0.00,0.00,1.00}{0.8676}
	& \textcolor[rgb]{0.00,0.00,1.00}{30.54} & \textcolor[rgb]{0.00,0.00,1.00}{0.8946}
	& \textcolor[rgb]{0.00,0.00,1.00}{27.80} & \textcolor[rgb]{0.00,0.00,1.00}{0.8464} 
	& \textcolor[rgb]{0.00,0.00,1.00}{30.75} & \textcolor[rgb]{0.00,0.00,1.00}{0.9074}\\
	IKC (ours) &  
	& \textcolor[rgb]{1.00,0.00,0.00}{36.62} & \textcolor[rgb]{1.00,0.00,0.00}{0.9658}
	& \textcolor[rgb]{1.00,0.00,0.00}{32.82} & \textcolor[rgb]{1.00,0.00,0.00}{0.8999}
	& \textcolor[rgb]{1.00,0.00,0.00}{31.36} & \textcolor[rgb]{1.00,0.00,0.00}{0.9097} 
	& \textcolor[rgb]{1.00,0.00,0.00}{30.36} & \textcolor[rgb]{1.00,0.00,0.00}{0.8949}
	& \textcolor[rgb]{1.00,0.00,0.00}{36.06} & \textcolor[rgb]{1.00,0.00,0.00}{0.9474}\\
	\hline
	Bicubic & \multirow{7}{*}{$\times 3$} 
	& 26.21 & 0.7766
	& 24.01 & 0.6662 
	& 24.25 & 0.6356 
	& 21.39 & 0.6203 
	& 22.98 & 0.7576\\
	CARN$^*$ \cite{CARN} &
	& 27.26 & 0.7855 
	& 25.06 & 0.6676 
	& 25.85 & 0.6566 
	& 22.67 & 0.6323 
	& 23.84 & 0.7620\\
	ZSSR \cite{ZSSR} &  
	& 28.25 & 0.7989
	& 26.11 & 0.6942 
	& 26.06 & 0.6633 
	& 23.26 & 0.6534 
	& 25.19 & 0.7914\\
	Pan \textit{et al.} \cite{Deblur} $+$ CARN \cite{CARN} &  
	& 19.05 & 0.5226 
	& 17.61 & 0.4558 
	& 20.51 & 0.5331 
	& 16.72 & 0.4578 
	& 18.38 & 0.6118\\
	CARN \cite{CARN} $+$ Pan \textit{et al.} \cite{Deblur} &  
	& 30.13 & 0.8562 
	& 27.57 & 0.7531 
	& 27.14 & 0.7152
	& 24.45 & 0.7241 
	& \B{27.67} & \B{0.8592}\\
	$\mathcal{P}+$ SFTMD &  
	& \textcolor[rgb]{0.00,0.00,1.00}{31.26} & \textcolor[rgb]{0.00,0.00,1.00}{0.9291} 
	& \textcolor[rgb]{0.00,0.00,1.00}{28.41} & \textcolor[rgb]{0.00,0.00,1.00}{0.7811}
	& \textcolor[rgb]{0.00,0.00,1.00}{27.37} & \textcolor[rgb]{0.00,0.00,1.00}{0.8102} 
	& \textcolor[rgb]{0.00,0.00,1.00}{24.57} & \textcolor[rgb]{0.00,0.00,1.00}{0.7458} 
	& 26.29 & 0.8399\\
	IKC (ours) &  
	& \textcolor[rgb]{1.00,0.00,0.00}{32.16} & \textcolor[rgb]{1.00,0.00,0.00}{0.9420}
	& \textcolor[rgb]{1.00,0.00,0.00}{29.46} & \textcolor[rgb]{1.00,0.00,0.00}{0.8229}
	& \textcolor[rgb]{1.00,0.00,0.00}{28.56} & \textcolor[rgb]{1.00,0.00,0.00}{0.8493}
	& \textcolor[rgb]{1.00,0.00,0.00}{25.94} & \textcolor[rgb]{1.00,0.00,0.00}{0.8165}
	& \textcolor[rgb]{1.00,0.00,0.00}{28.21} & \textcolor[rgb]{1.00,0.00,0.00}{0.8739}\\
	\hline
	Bicubic & \multirow{7}{*}{$\times 4$} 
	& 24.57 & 0.7108 
	& 22.79 & 0.6032 
	& 23.29 & 0.5786 
	& 20.35 & 0.5532 
	& 21.50 & 0.6933\\
	CARN$^*$ \cite{CARN} &
	& 26.57 & 0.7420 
	& 24.62 & 0.6226 
	& 24.79 & 0.5963 
	& 22.17 & 0.5865 
	& 21.85 & 0.6834\\
	ZSSR \cite{ZSSR} &
	& 26.45 & 0.7279 
	& 24.78 & 0.6268 
	& 24.97 & 0.5989 
	& 22.11 & 0.5805 
	& 23.53 & 0.7240 \\
	Pan \textit{et al.} \cite{Deblur} $+$ CARN \cite{CARN} &  
	& 18.10 & 0.4843 
	& 16.59 & 0.3994 
	& 18.46 & 0.4481
	& 15.47 & 0.3872
	& 16.78 & 0.5371\\
	CARN \cite{CARN} $+$ Pan \textit{et al.} \cite{Deblur} &  
	& 28.69 & 0.8092 
	& 26.40 & 0.6926
	& 26.10 & 0.6528 
	& \B{23.46} & 0.6597 
	& \B{25.84} & \B{0.8035}\\
	$\mathcal{P}+$ SFTMD &  
	& \textcolor[rgb]{0.00,0.00,1.00}{29.29} & \textcolor[rgb]{0.00,0.00,1.00}{0.9014} 
	& \textcolor[rgb]{0.00,0.00,1.00}{26.40} & \textcolor[rgb]{0.00,0.00,1.00}{0.7137} 
	& \textcolor[rgb]{0.00,0.00,1.00}{26.16} & \textcolor[rgb]{0.00,0.00,1.00}{0.7648} 
	& 22.97 & \B{0.6722}
	& 24.24 & 0.7950\\
	IKC (ours) &  
	& \textcolor[rgb]{1.00,0.00,0.00}{31.52} & \textcolor[rgb]{1.00,0.00,0.00}{0.9278} 
	& \textcolor[rgb]{1.00,0.00,0.00}{28.26} & \textcolor[rgb]{1.00,0.00,0.00}{0.7688} 
	& \textcolor[rgb]{1.00,0.00,0.00}{27.29} & \textcolor[rgb]{1.00,0.00,0.00}{0.8014} 
	& \textcolor[rgb]{1.00,0.00,0.00}{25.33} & \textcolor[rgb]{1.00,0.00,0.00}{0.7760} 
	& \textcolor[rgb]{1.00,0.00,0.00}{29.90} & \textcolor[rgb]{1.00,0.00,0.00}{0.8793}\\
	\hline
	\end{tabular}
	\end{center}\vspace{-0.6cm}
\end{table*}
\subsection{Experiments of SFTMD}
We evaluate the performance of the proposed SFTMD on different Gaussian kernels.
The kernel settings are given in \tablename~\ref{tab:SFT}.
We compare the SFTMD with the SOTA non-blind SR methods SRCNN-CAB \cite{CAB} and SRMD \cite{SRMD}.
As SFTMD adopts SRResNet as the main network, which is different from SRMD and SRCNN-CAB, we provide two additional baselines that have same network structures but different concatenation strategies:
(1) SRResNet with concatenating $\mathcal{H}$ at the first layer,
(2) SFTMD with SFT layer replaced by direct concatenation\footnote{Direct concatenation means concatenating the kernel maps with feature maps directly. This is different from the affine transformation in the SFT layer.}.
\tablename~\ref{tab:SFT} shows the quantitative comparison results.
Comparing with the SOTA SR methods -- SRCNN-CAB and SRMD, the proposed SFTMD achieves significantly better performance on all settings and dataset.
Comparing with two additional baselines that all use SRResNet as the main network, SFTMD could also obtain the best results.
This further demonstrated the effect of SFT layers.
It is worth noting that directly concatenating $\mathcal{H}$ in SRResNet will cause severe performance drop.
As the combination of direct concatenation strategy and residual structure will interfere with image processing and cause severe artifacts.
\subsection{Experiments on Synthetic Test Images}
\begin{figure*}[!t]\vspace{-0.2cm}
	\scriptsize{
	\begin{center}
		\subfigure[LR image]{\includegraphics[width=0.19\linewidth]{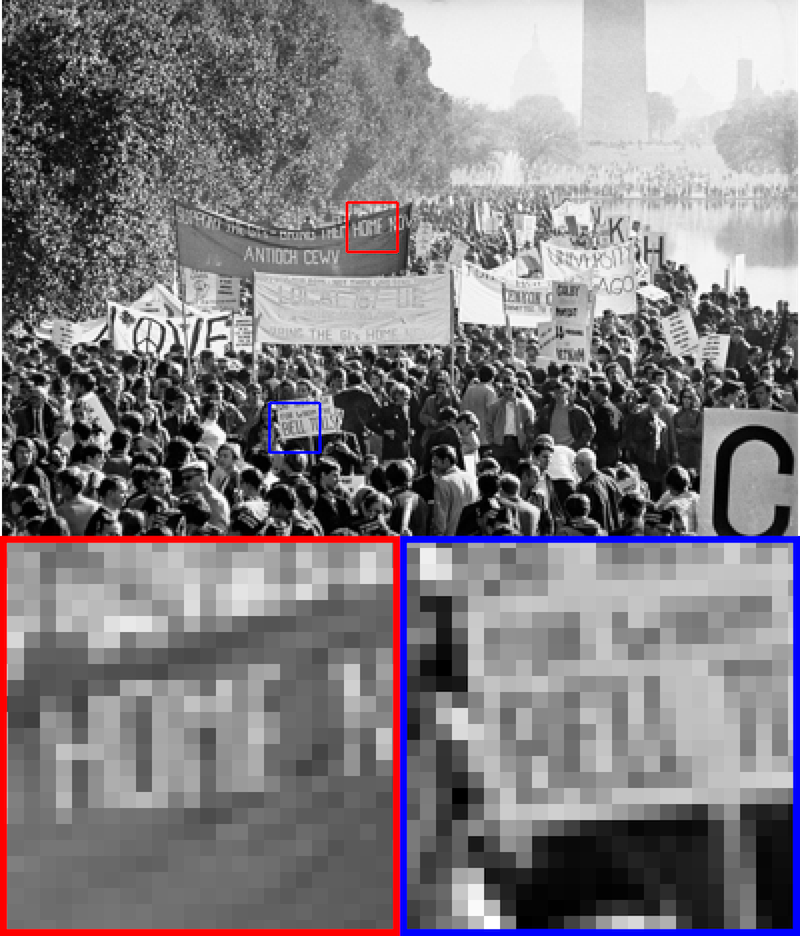}}
		\hspace{0.00cm}
		\subfigure[A+ \cite{A+}]{\includegraphics[width=0.19\linewidth]{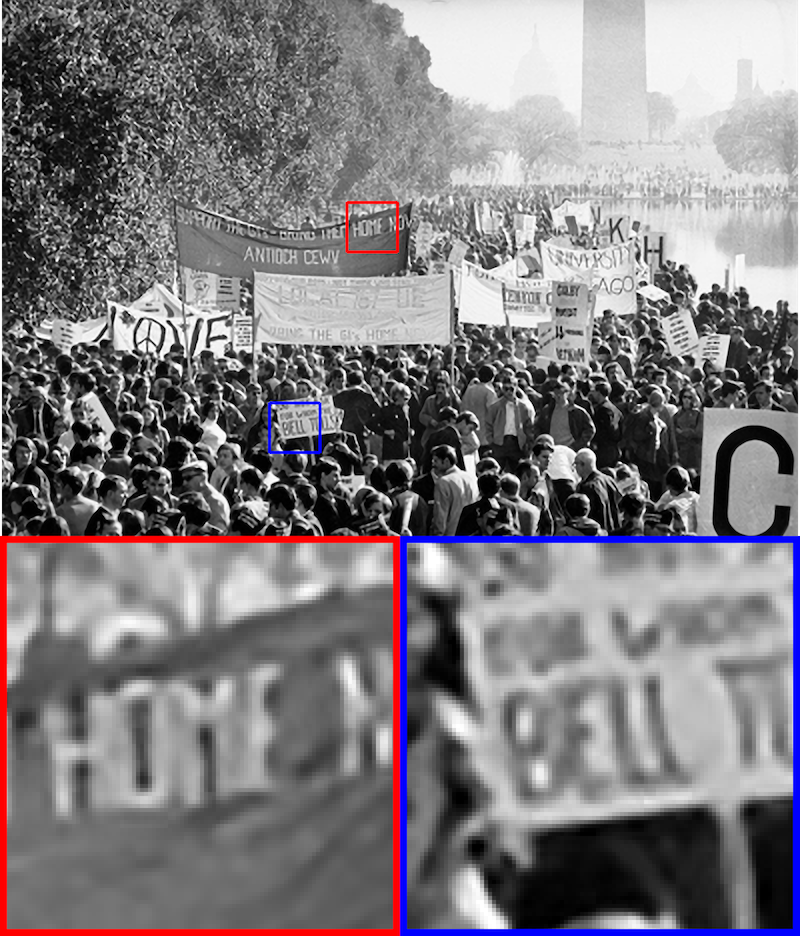}}
		\hspace{0.00cm}
		\subfigure[ZSSR \cite{ZSSR}]{\includegraphics[width=0.19\linewidth]{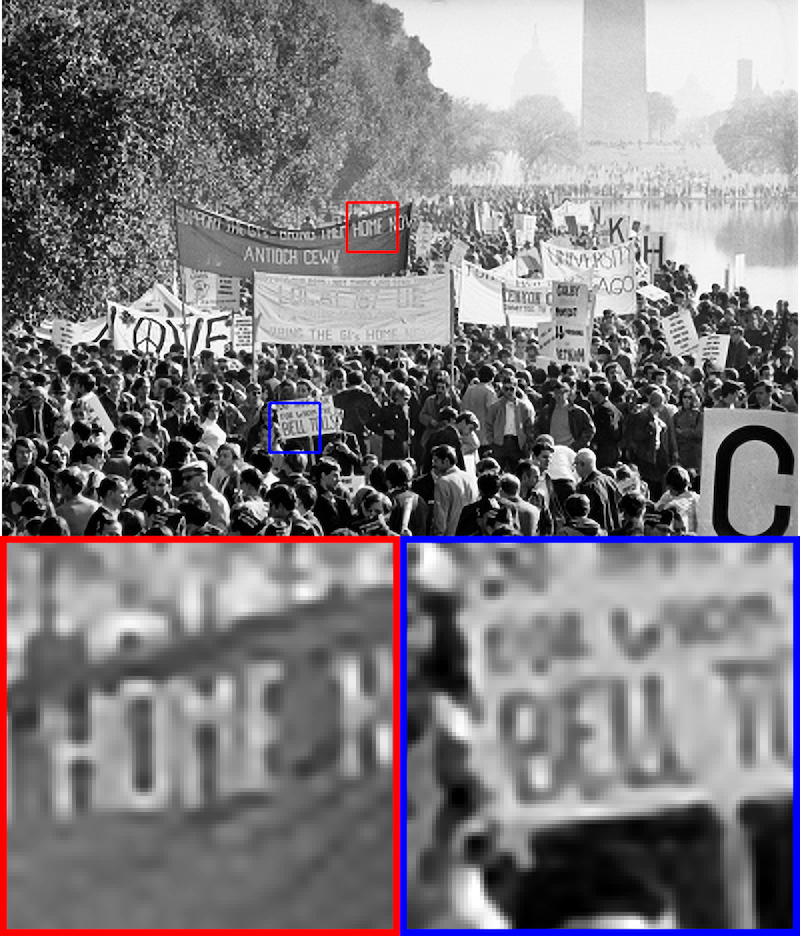}}
		\hspace{0.00cm}
		\subfigure[CARN \cite{CARN}]{\includegraphics[width=0.19\linewidth]{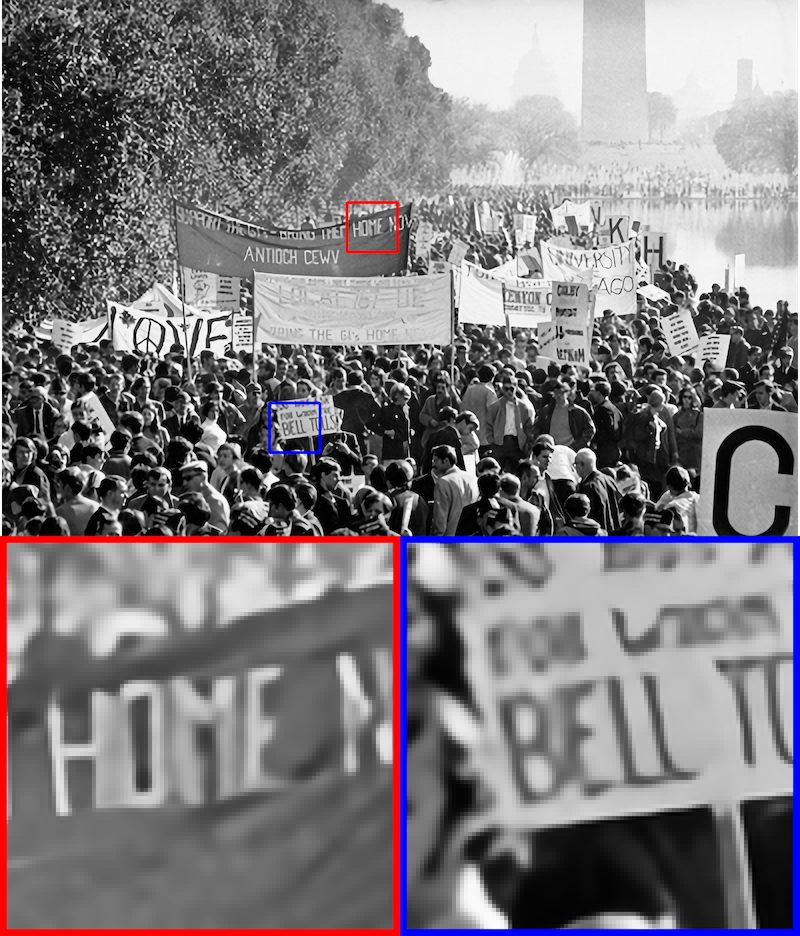}}
		\hspace{0.00cm}
		\subfigure[IKC (ours)]{\includegraphics[width=0.19\textwidth]{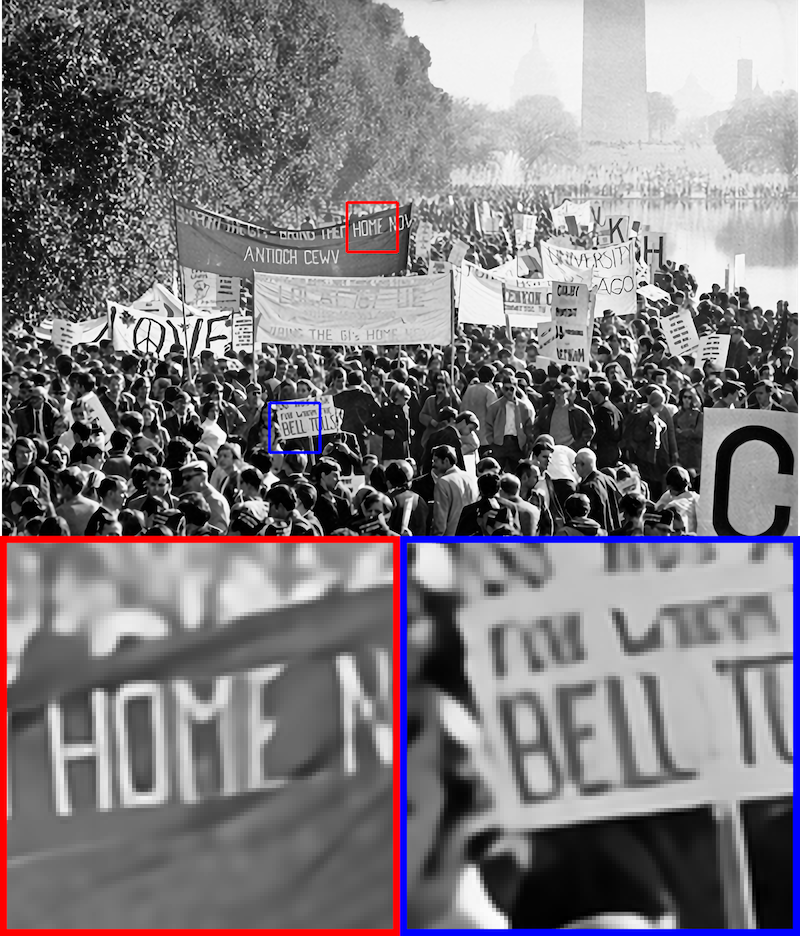}}
		\caption{SISR performance comparison of different methods with SR factor 4 on a real historic image `\emph{1967 Vietnam war protest}'.}
		\label{fig:history}
	\end{center}}\vspace{-0.66cm}
\end{figure*}
We evaluate the performance of the proposed method on the synthetic test images.
\figurename~\ref{fig:stepcom} shows the intermediate results during correction.
As one can see that the SR results using the kernel estimated by the predictor directly (the initial prediction in \figurename~\ref{fig:stepcom}) are unsatisfactory and contain either blurry or ringing artifacts.
As the number of iterations increases, artifacts and blurring are gradually alleviated.
The quantitative results (PSNR) also prove the necessity of the iterative correction strategy.
We can see at the $4$th iteration, the SR results using corrected kernels are able to show good visual quality.

\begin{figure}[!t]
	\begin{center}\vspace{-0.2cm}
	\includegraphics[width=1.0\linewidth]{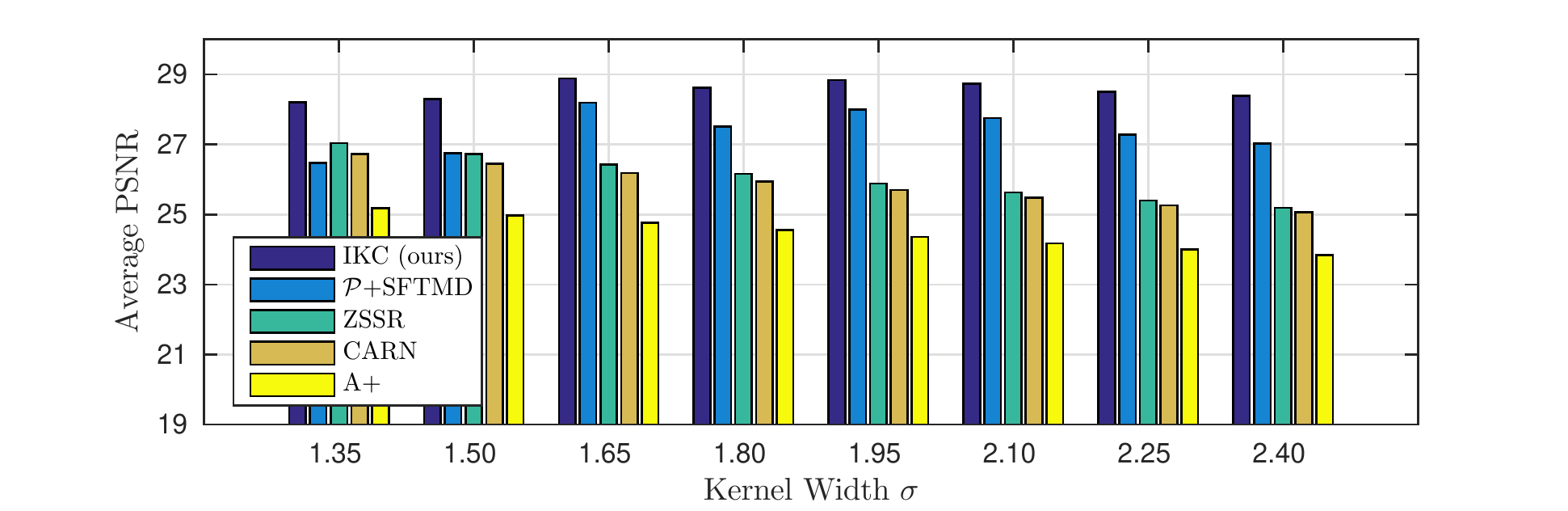}
	\caption{The PSNR performance of different methods on BSD100 \cite{BSD100} with different kernel width. The test SR factor is $3$.}
	\label{fig:kernelcom}
	\end{center}\vspace{-0.8cm}
\end{figure}

We then conduct thorough comparisons with the SOTA non-blind and blind SR methods using \textit{Gaussian8} kernels.
We also provide the comparison with the solutions using the SOTA deblurring method.
We perform blind debluring method Pan \textit{et al.} \cite{Deblur} before and after the non-blind SR method CARN \cite{CARN}.
\tablename~\ref{tab:Gaus8} shows the PSNR and SSIM \cite{SSIM} results on five widely-used datasets.
As one can see, despite the remarkable performance under bicubic downsampling setting, the non-blind SR methods suffer severe performance drop when the downsampling kernel is different from the predefined bicubic kernel.
The ZSSR \cite{ZSSR} takes the effect of blur kernel into account, and provides better SR performance compared with non-blind SR methods.
Performing blind deblurring on the LR images makes the SR images sharper, but lost in image quality
The final SR results have severe distortion.
Deblurring on the blurred super-resolved images provides better results, but fails to reconstruct textures and details.
Although the SR results without kernel correction (denoted by ``$\mathcal{P}+$SFTMD'') achieves comparable quantitative performance with the existing methods, the SR performance can still be greatly improved by using the proposed IKC method.
An example is shown in \figurename~\ref{fig:Gaus}.
The PSNR values of different methods on different blur kernels are shown in \figurename~\ref{fig:kernelcom}.
As can be seen, when the kernel width becomes larger, the SR performance of the previous methods decreases.
Meanwhile, the proposed IKC method achieves superior performance under all blur kernels.
\begin{table}[!t]
	\scriptsize
	\begin{center}
	\caption{Quantitative performance of the proposed IKC method on other downsampling settings.}
	\label{tab:nonGaus}
	\vspace{0.8mm}
	\begin{tabular}{lccccc}
	\hline
	\multirow{2}{*}{Method} & Kernel & \multicolumn{2}{c}{BSD100 \cite{BSD100}} & \multicolumn{2}{c}{BSD100 \cite{BSD100}}\\
	& Width & PSNR & SSIM & PSNR & SSIM\\
	\hline
	\hline
	CARN \cite{CARN} & \multirow{6}{*}{2.0} &
	26.05 & 0.6970 & 25.92 & 0.6601 \\
	ZSSR \cite{ZSSR} & &
	25.64 & 0.6771 & 25.64 & 0.6446 \\
	CARN \cite{CARN}$+$Pan \textit{et al.} \cite{Deblur} & &
	25.71 & 0.7115 & 25.94 & 0.6804 \\
	$\mathcal{P}+$ SFTMD & &
	23.42 & 0.6812 & 25.01 & 0.7231 \\
	IKC, \textit{w/o} PCA & &
	\B{26.85} & \B{0.7694} & \B{26.30} & \B{0.7812} \\
	IKC (ours) & &
	\R{27.06} & \R{0.7704} & \R{26.35} & \R{0.7838} \\
	\hline
	CARN \cite{CARN} & \multirow{6}{*}{3.0} &
	24.20 & 0.6066 & 24.53 & 0.5812 \\
	ZSSR \cite{ZSSR} & &
	24.19 & 0.6045 & 24.53 & 0.5796 \\
	CARN \cite{CARN}$+$Pan \textit{et al.} \cite{Deblur} & &
	25.62 & 0.6678 & 25.52 & 0.6293 \\
	$\mathcal{P}+$ SFTMD & &
	23.30 & 0.6799 & 24.41 & 0.7214 \\
	IKC, \textit{w/o} PCA & &
	\B{26.75} & \B{0.7685} & \B{26.28} & \B{0.7849} \\
	IKC (ours) & &
	\R{26.98} & \R{0.7694} & \R{26.58} & \R{0.7994} \\
	\hline
	\end{tabular}
	\end{center}\vspace{-0.8cm}
\end{table}

To further show the generalization ability of the proposed IKC method, we test our method on another widely-used degradation setting \cite{Benckmark}, which involves Gaussian kernels and direct downsampler.
When the downsampling function is different, the LR images obtained by the same blur kernel are also different.
\tablename~\ref{tab:nonGaus} shows the quantitative results of the proposed IKC method under different downsampling settings.
The proposed IKC method has maintained its performance, which indicates that IKC is able to generalize to a downsampling setting that is inconsistent with the training settings.
An important reason why the IKC method has such generalization ability is that IKC learns the kernel after PCA rather than the kernel parameterized by kernel width.
PCA provides a feature representation for the kernels.
IKC learns the relationship between the SR images and these features rather than the Gaussian kernel width.
In \tablename~\ref{tab:nonGaus}, we provide the comparison with the IKC method that adopts kernels parameterized by Gaussian kernel width.
Experiments prove that the use of PCA helps to improve the generalization performance of IKC. 
\subsection{Experiments on Real Images Set}
Besides the above experiments on synthetic test images, we also conduct experiments on real images to demonstrate the effectiveness of the proposed IKC and SFTMD.
Since there are no ground-truth HR images, we only provide the visual comparison.
\figurename~\ref{fig:history} shows the SISR results on real world image from the Historic dataset.
For comparison, the A+ \cite{A+} and CARN \cite{CARN} are used as the representative SR methods with bicubic downsampling, and ZSSR \cite{ZSSR} is used as the representative blind SR method.
For a real-world image, the downsampling kernel is unknown and complicated, thus performance of the non-blind SR methods are severely affected.
The SOTA blind method -- ZSSR also fails to provide satisfactory results.
In comparison, IKC provides artifact-free SR result with sharp edges.

\begin{figure}[!t]
	\begin{center}\vspace{-1mm}
		\subfigure[\scriptsize LR image]{\includegraphics[width=0.234\textwidth]{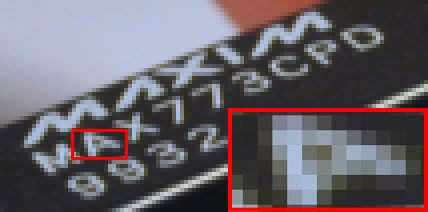}}
		\hspace{0.00cm}
		\subfigure[\scriptsize ZSSR \cite{ZSSR}]{\includegraphics[width=0.234\textwidth]{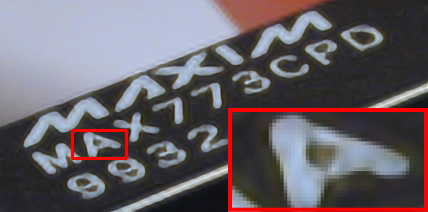}}
		
		\vspace{-0.23cm}
		\subfigure[\scriptsize SRMD with hand-craft kernel]{\includegraphics[width=0.234\textwidth]{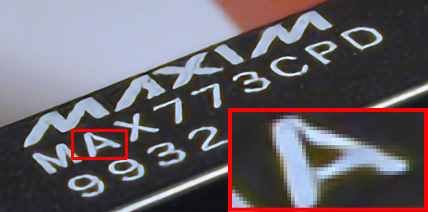}}
		\hspace{0.00cm}
		\subfigure[\scriptsize IKC (Ours)]{\includegraphics[width=0.234\textwidth]{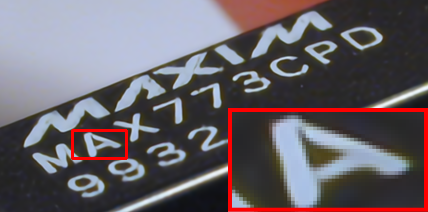}}
		\caption{SR results of the real image ``\emph{Chip}'' with SR factor 4. The hand-craft kernel width suggested by SRMD is 1.5.}
		\label{fig:hand}
	\end{center}\vspace{-0.8cm}
\end{figure}

We also compare the proposed IKC method with the non-blind SR method using `hand-craft' kernel on real-world image `\emph{Chip}'.
We super-resolve the LR image using SRMD with the `hand-craft' kernel suggested by \cite{SRMD}. 
They use a grid search strategy to find the kernel parameters with good visual quality.
The visual comparison is shown in \figurename~\ref{fig:hand}.
We can see that the result of SRMD has harper edges and higher contrast, but also looks a little artificial.
At the same time, IKC could provide visual pleasing SR results \emph{automatically}.
Although the contrast of IKC result is not as high as SRMD result, it still provides sharp edges and more natural visual effects.
\section{Discussion}
In this paper, we explore the relationship between blur kernel mismatch and the SR results, then propose an iterative blind SR method -- IKC.
We also propose SFTMD, a new SR network architecture for multiple blur kernels.
In this paper, our experiments are mainly conducted on the isotropic kernels.
However, the isotropic kernels don't seem to be applicable in some real world applications.
As in most cases, there are some slightly motion blurs that affect the kernel.
It is worth noting that the asymmetry of the kernel mismatch effect that IKC relies on can still be observed in the case of slightly motion blur (anisotropy blur kernels).
For example, the artifacts and blur of a SR image in a certain direction is related to the width of the kernel in the same direction.
This indicates that, by employing such asymmetry of the kernel mismatch in each direction, the IKC method can also be applied to more realistic cases with slightly motion blur, which will be our future work.

\textbf{Acknowledgements.}
This work is partially supported by SenseTime Group Limited, National Key Research and Development Program of China (2016YFC1400704), Shenzhen Research Program (JCYJ20170818164704758, JCYJ20150925163005055, CXB201104220032A), and Joint Lab of CAS-HK.

\clearpage
{\small
\bibliographystyle{ieee_fullname}
\bibliography{Blind}
}

\end{document}